\documentclass{article}

\usepackage[numbers]{natbib}

\usepackage[final]{neurips_2020}

\usepackage[utf8]{inputenc} 
\usepackage[T1]{fontenc}    
\usepackage{hyperref}       
\usepackage{url}            
\usepackage{booktabs}       
\usepackage{amsfonts}       
\usepackage{nicefrac}       
\usepackage{microtype}      
\usepackage[pdftex]{graphicx}
\usepackage[toc,page]{appendix}
\usepackage{amsmath}

\title{Physics-Informed Neural Network Super Resolution for Advection-Diffusion Models}

\author{
Chulin Wang; Eloisa Bentivegna; Wang Zhou; Levente J. Klein; Bruce Elmegreen \\
IBM Research\\
\{wangc,eloisa.bentivegna,wang.zhou\}@ibm.com; \{kleinl,bge\}@us.ibm.com \\
}

\begin{document}

\maketitle

\begin{abstract}
Physics-informed neural networks (NN) are an emerging technique to improve spatial resolution and enforce physical consistency of data from physics models or satellite observations. A super-resolution (SR) technique is explored to reconstruct high-resolution images (4x) from lower resolution images in an advection-diffusion model of atmospheric pollution plumes. SR performance is generally increased when the advection-diffusion equation constrains the NN in addition to conventional pixel-based constraints. The ability of SR techniques to also reconstruct missing data is investigated by randomly removing image pixels from the simulations and allowing the system to learn the content of missing data. Improvements in S/N of 11\% are demonstrated when physics equations are included in  SR with 40\% pixel loss. Physics-informed NNs accurately reconstruct corrupted images and generate better results compared to the standard SR approaches.
\end{abstract}

\section{Introduction}

Modeling physical systems is often limited to coarse spatial and temporal grid resolution due to the exponential dependence of computing requirements on the grid sizes \cite{earthcomp}. While traditional super resolution (SR) techniques \cite{dong2014learning, Leidig:2017gan, lim2017enhanced,  wang2018esrgan} can boost model granularity by minimizing pixel-level differences between high-resolution (HR) data and super-resolved output made from low resolution (LR) input, the output may not capture the physical, ecological or geological processes at work that are governed by physical laws. Besides, real observations like satellite images \cite{sentinel5p} are sparse and often incomplete, leading to ``missing pixels.'' How to fill in missing values while maintaining physical consistency remains an open question.

Here we propose a physics-informed neural network for SR (PINNSR) method that incorporates both traditional SR techniques and fundamental physics. In addition to minimizing pixel-wise differences, PINNSR also enforces the governing physics laws by minimizing a physics consistency loss.
We apply PINNSR to plume simulations based on the advection-diffusion equation with variable wind conditions, which
is considered as a proxy for remote satellite observations of pollutant gas dispersion \cite{sentinel5p}. Compared to traditional SR methods, our approach demonstrates that:
\begin{itemize}
    \item Data from a first-principle advection-diffusion equation at low resolution can be forced to reconstruct physically meaningful data rather than the numerical interpolation of the LR data.
    \item The additional physics constraints increase the accuracy in reconstruction of HR data for both physics-governed processes and missing-value conditions.
    \item Physics consistency loss can quantify how reliably the SR generated data reproduce the physics laws. 
    \item SR for physics-related data should be modeled from direct observations of LR and HR data, compared to synthetic LR from bicubic downsampling in typical computer vision problems. 

\end{itemize}

\section{Related Work}

SR with neural networks (NN) has been extensively studied in recent years.
SRCNN \cite{dong2014learning} firstly adopted a three-layer CNN to represent the mapping function. Deeper and wider networks \cite{kim2016, kim2016recursive, dai2019, tai2017, tai2017memnet} with residual learning were proposed to enhance the performance. 
SRGAN \cite{Leidig:2017gan} adopted generative adversarial networks (GANs) \cite{gan} and showed better perceptual quality. EDSR \cite{lim2017enhanced} improved the generation by removing batch-normalization, and residual-in-residual blocks (RRDB) introduced by ESRGAN \cite{wang2018esrgan} further boosted the performance.

There has been growing interest in applying SR to physics-related data. Fukami \emph{et al.} \cite{fukami2018super} used the SRCNN network structure to super-resolve 2D laminar cylinder flow. MeshfreeFlowNet \cite{jiang2020meshfreeflownet} used a U-Net structure to reconstruct the Rayleigh-B\'enard instability. PIESRGAN \cite{bode2019using} utilized the ESRGAN architecture for turbulence modeling. In all of these models, the LR input was generated from down-sampling the HR dataset, so part of the NN learning could be the reverse of the down-sampling algorithm itself, making the models unpractical for data deviating from the same down-sampling process. 




\section{Method}

\subsection{Plume simulations}
\label{sec:sims}

Common prior arts for SR use down-sampled HR images to approximate the LR input.
This assumption limits the modeling to be the down-sampling kernel (normally bicubic interpolation). 
For example, unstable flows like the Rayleigh-Taylor instability, which grows fastest at the shortest wavelengths, have different growth rates and structures at different resolutions. Thus a naive bicubic interpolation cannot capture the mapping between LR and HR.

Here we simulate atmospheric dispersion of gaseous plumes through  
integration of the advection-diffusion equation for the gas concentration $C$ in 2D:
\begin{equation}
    \partial_t C + \nabla \cdot C {\bf u} = \nabla \cdot ({\bf K} \cdot \nabla C) + S,
    \label{eq:ade}
\end{equation}
where ${\bf u}$ is the atmospheric velocity field, ${\bf K}$ is the diffusivity tensor, and $S$ is a source (or sink) term. 
In this work, LR and HR datasets are \emph{both} generated by running the simulation model twice but at different spatial grid sizes for each of the random source placements, with the HR simulation having $4\times$ finer resolution than the LR simulation. 
Snapshots of the gas concentration $C^{(t)}$ are saved to construct LR-HR (input-output) training pairs.
More details can be found in Appendix~\ref{sec:app}.




\subsection{PINNSR network}

\begin{figure}[htbp]
    \centering
    \includegraphics[width=0.95\linewidth]{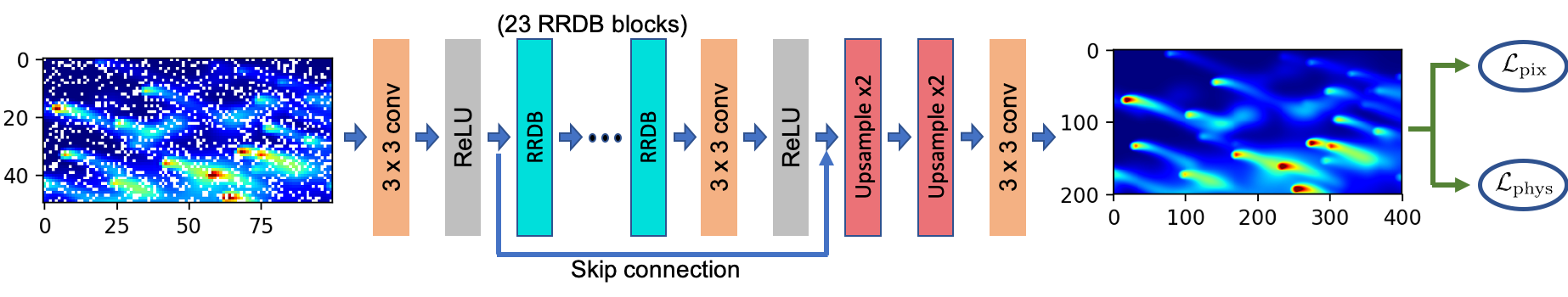}
    \vspace{-2mm}
    \caption{
        Network structure of PINNSR. The input LR is generated by simulating on a coarse grid instead of down-sampling from HR. Random pixels are dropped (shown in white) to imitate missing pixels. The total loss is a weighted sum of pixel loss $\mathcal{L}_{\rm pix}$ and physics consistency loss $\mathcal{L}_{\rm phys}$.
    }
    \label{fig:network}
\end{figure}

\begin{figure}[htbp]
    \centering
    \vspace{-0.1in}
    \includegraphics[width=0.9\linewidth]{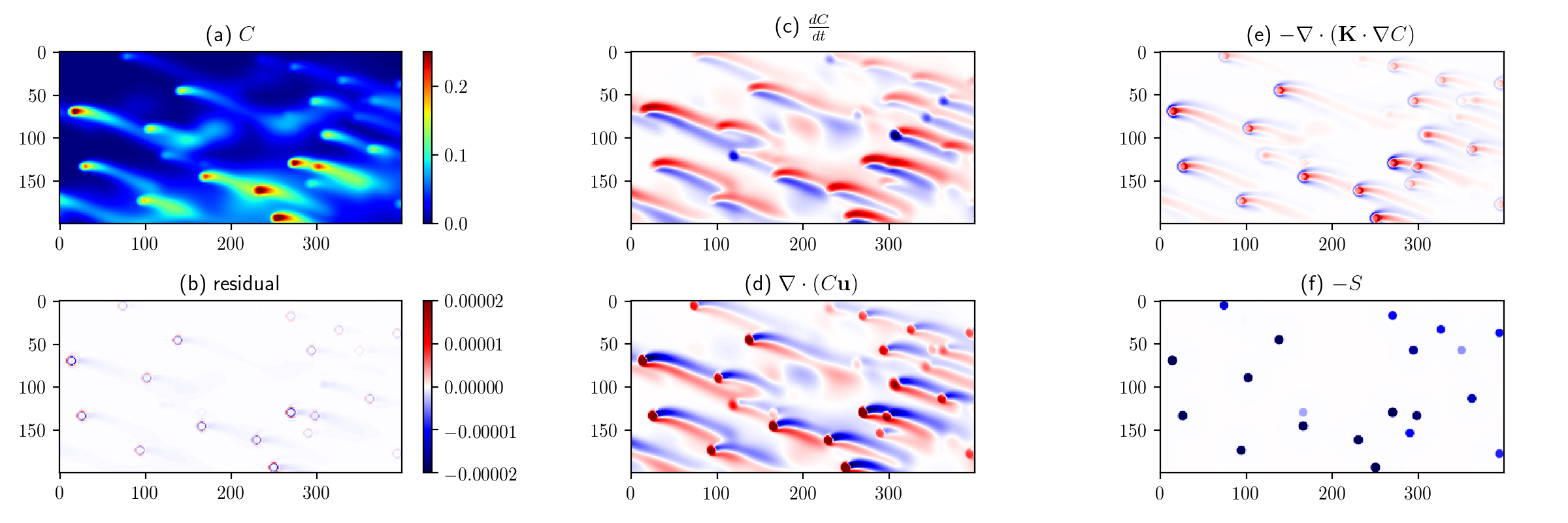}
    \vspace{-0.1in}
    \caption{(a) The HR image for 20 randomly distributed plume sources under variable wind conditions; (b) Physics residual term $R(C)$ due to numerical rounding errors; (c) - (f) Different terms of advection-diffusion equation that contribute to the HR image.
    }
    \label{fig:pde}
    \vspace{-0.05in}
\end{figure}

The base network of PINNSR is built on multiple RRDB blocks \cite{wang2018esrgan}, as shown in Figure~\ref{fig:network}. Whereas \cite{wang2018esrgan} employs a discriminator to improve the visual quality at a cost of reduced peak signal-to-noise ratio (PSNR), for physics-based data, it is preferred to have high PSNR. Thus, no discriminator is included for PINNSR. Instead, we introduce a \emph{physics consistency loss}
\begin{equation}
    \mathcal{L}_{\rm phys} = ||R(C_{\rm SR}) - R(C_{\rm HR})||_1,
    \label{eq:phys_loss}
\end{equation}
which minimizes the physics residual $R(C)$ between SR and HR.
The physics residual $R(C)$ is defined from the governing advection-diffusion equation:
\begin{equation}
    R(C) = \partial_t C + \nabla \cdot C {\bf u} -\nabla \cdot ({\bf K} \cdot \nabla C) - S,
    \label{eq:RC}
\end{equation}
where the derivatives are calculated using a finite-difference approximation.
Due to this approximation and the resulting truncation error, $R(C)$ is not zero but the sum of all the higher-order terms neglected when computed at the relevant resolution, and thus Equation~\ref{eq:phys_loss} minimizes the difference in $R(C)$.
A visualization of the HR image and each term of the corresponding physics residual is illustrated in Figure \ref{fig:pde}. As shown in panel (b), the residual is mostly 0 for the entire image, except near the ``edge'' of each source location and along the center of the plume where the truncation error is the highest. 


As depicted in Figure~\ref{fig:network}, the total loss $\mathcal{L}_{\rm tot}$ is a weighted sum of the pixel loss $\mathcal{L}_{\rm pix}$ and the physics consistency loss $\mathcal{L}_{\rm phys}$, with weighting parameter $\eta$ and batch size $N$:
\begin{equation}
    \begin{aligned}
    \mathcal{L}_{\rm tot} & = \mathcal{L}_{\rm pix} + \eta \cdot \mathcal{L}_{\rm phys}\\
     & = 
     \frac{1}{N} \sum_{i=1}^{N}{
        ||C_{\rm SR} - C_{\rm HR}||_1
    }
     + 
    \eta \cdot 
    \frac{1}{N} \sum_{i=1}^{N}{
        ||R(C_{\rm SR}) - R(C_{\rm HR})||_1.
    }
    \end{aligned}
    \label{eq:tot_loss}
\end{equation}

\section{Experiments}

\subsection{Dataset}

As explained in Section~\ref{sec:sims}, the dataset is constructed to simulate the atmospheric dispersion of gaseous plumes using Equation~\ref{eq:ade} for both LR and HR spatial scales. For each of the input-output pairs, snapshots of gas concentration are stacked as 3-channel images in the order of $C^{(t-1)}$, $C^{(t)}$, and $C^{(t+1)}$ to allow estimation of the time derivative. The spatial gradients and time derivatives for the physics equation are evaluated by first-order centered differences on the space-time grid.

To make the problem more physically relevant to real satellite data, some of the pixels in the LR images are randomly dropped to simulate cloud cover, non-convergent flux calculations, and other glitches. Experiments are conducted at various dropping rates (0\%, 20\%, 40\%, and 60\%) to study the robustness of the method. 

\subsection{Baselines and metrics}

{\bf Bicubic.} 
For a baseline model, we use bicubic interpolation to first fill the missing pixels (if there are any), and then the SR images are generated by a $4\times$ bicubic upsampling.

{\bf Downsampled HR (Dwn-HR).} The network is trained with downsampled HR as input. This is commonly done for SR datasets \cite{div2k, flickr2k, urban100, wang2018esrgan}. However, to mimic the real situation where LR is available but not HR, we test the trained model on \emph{simulated} LR instead of downsampled HR.

{\bf Standard SR (Std-SR).} The network is trained on simulated LR and HR pairs but without the physics consistency loss $\mathcal{L}_{\rm phys}$.

{\bf PINNSR.} Our proposed approach trains the network on simulated LR and HR pairs with a weighted sum of the pixel loss $\mathcal{L}_{\rm pix}$ and the physics consistency loss $\mathcal{L}_{\rm phys}$.

{\bf Metrics.} Models are evaluated by the standard PSNR and structural similarity (SSIM). In addition, we include the physics consistency loss $\mathcal{L}_{\rm phys}$ as another metric to measure the fidelity on governing physics laws.
Visual illustrations of the pixel differences are also presented.

\subsection{Results}

\begin{table}[htbp]
  \caption{Comparison of test results for different models (the best results are in bold).}
  \label{table:compare}
  \centering
  \scalebox{0.8}{
  \begin{tabular}{ccccc}
    \toprule
    & \multicolumn{4}{c}{$\rm PSNR$ / $1 - \rm SSIM$ / $\mathcal{L}_{\rm phys}$ } \\
    \cmidrule(r){2-5}
    Pixel drop & Bicubic & Dwn-HR & Std-SR  & PINNSR   \\
    \midrule
    0\%     & 45.73 / 0.0065 / 5.9E-7 & 
              45.85 / 0.0058 / 1.1E-6 &
              82.29 / 2.1E-6 / 2.8E-7 & 
              \textbf{82.83} / \textbf{2.0E-6} / \textbf{0.9E-7} \\
              
    20\%    & 45.42 / 0.0070 / 6.4E-7 & 
              45.17 / 0.0064 / 1.5E-6 & 
              82.35 / \textbf{0.9E-6} / 1.8E-7 & 
              \textbf{82.71} / 1.5E-6 / \textbf{1.2E-7} \\
              
    40\%    & 44.79 / 0.0080 / 7.2E-7 &
              44.83 / 0.0065 / 1.1E-6 & 
              81.19 / 1.5E-6 / 3.0E-7 &
              \textbf{82.12} / \textbf{1.3E-6} / \textbf{1.5E-7} \\
              
    60\%    & 43.26 / 0.0128 / 8.8E-7 & 
              44.66 / 0.0061 / 0.8E-6 & 
              78.44 / 2.7E-6 / 4.1E-7 & 
              \textbf{79.02} / \textbf{2.3E-7} / \textbf{2.6E-7} \\
    \bottomrule
  \end{tabular}
  }
\end{table}

\begin{figure}[htbp]
    \centering
    \includegraphics[width=\linewidth]{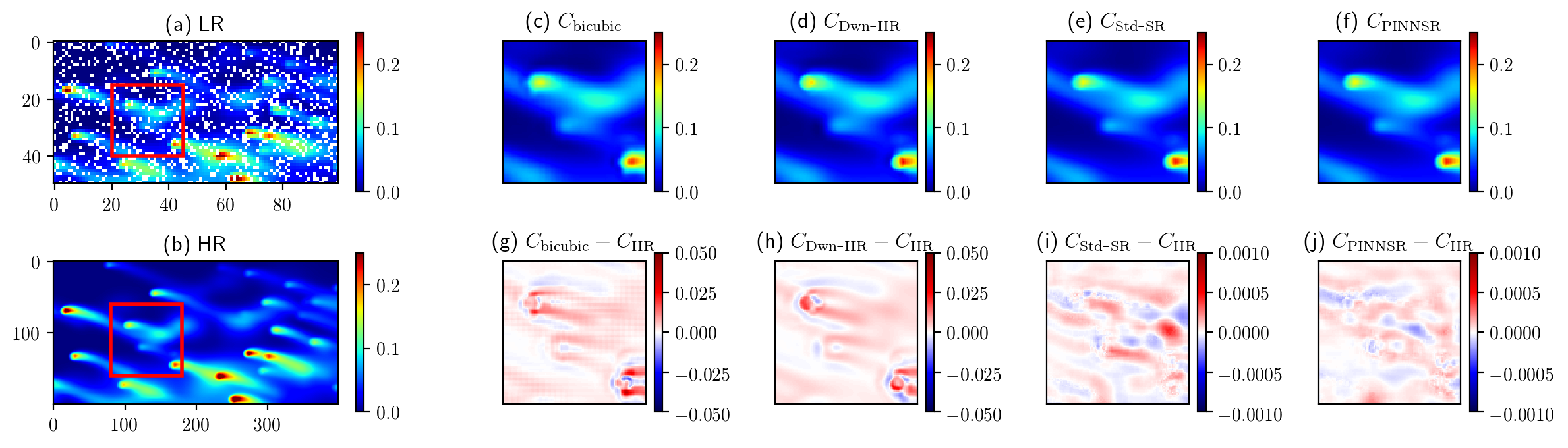}
    \caption{
        Qualitative comparison between different SR models from the test set when 20\% of the pixels are dropped. (a) LR input; (b) ground truth HR output; (c) - (f) SR generated by bicubic, Dwn-HR, Std-SR, and PINNSR; (g) - (j) the corresponding pixel residual calculated from $C_{\rm SR} - C_{\rm HR}$, the PINNSR clearly out-performs other models. Additional visualizations can be found in Appendix~\ref{sec:add_result}.
    }
    \label{fig:visual}
\end{figure}

Table~\ref{table:compare} summarizes our results. In all cases, PINNSR yields higher PSNR and better physics consistency, which confirms that the physics consistency loss introduced at training helps to regulate the learning and improve the performance.
The additional physics information not only enforces the output to comply with physics laws better (lower $\mathcal{L}_{\rm phys}$), but also improves the accuracy in generating SR output. As shown in Figure~\ref{fig:visual}(j), the pixel difference of PINNSR is smaller than Std-SR and 2 orders of magnitude lower compared to Bicubic. 
The improvement is persistent for all pixel drop rates, and the case with 40\% pixel drop has the best improvement with physics compared to Std-SR, increasing PSNR by 0.93 dB which corresponds to an 11\% decrease in rms error.

Dwn-HR can achieve unusually high PSNR $\sim 100$ (Appendix~\ref{sec:down-sample}) when the training and testing are both conducted on downsampled HR as input. But the same model, when tested with simulated LR, performs poorly with PSNR $\sim 45$, which is close to bicubic upsampling. Comparing Figure~\ref{fig:visual} (h) and (g), the patterns are very similar. This indicates that when trained with bicubically downsampled HR as input, Dwn-HR learns a reverse mapping (bicubic upsampling). When applied to data that differ from bicubic interpolation (like simulated LR), the performance of the model degrades significantly. Therefore, for physics-based data shown here, it is better to learn the mapping from LR to HR from direct observations of LR and HR rather than using downsampled HR with the assumption that the mapping can be captured by a known kernel (like bicubic interpolation).

To perform SR with missing pixels, an intuitive approach is to use bicubic interpolation to fill in the missing pixels first and then pass it to an SR model. We show that PINNSR learns the relations between existing and missing pixels and performs better than the two-step approach. More details are explained in Appendix~\ref{sec:bicubic_interp}.

\section{Conclusions}

We proposed a PINNSR method for super resolution on advection-diffusion modeling and demonstrated superior performance on both reconstruction accuracy and physics consistency. This is done by introducing a physics consistency loss to regulate model training. The method is robust even if pixels are missing as commonly observed in satellite images. The method can be generalized to other physics problems governed by different physics laws.

\clearpage

\section*{Broader Impact}

Multiple satellites are enabling remote observations of the Earth’s surface, even multiple times per day. Satellite images are affected by missing pixels and resolution that is spatially too coarse to identify greenhouse gas sources and polluters. Here we introduce a Physics-Informed Neural Network that creates physically consistent high-resolution imagery from low quality and low-resolution simulations based on an advection-diffusion equation. The reconstructed missing data follows the underlying physics law and demonstrates a robust way to ensure the physics consistency of the super-resolution imagery. Reconstructing missing data and increasing the spatial resolution of current satellite imagery, based on solid physics principles, can create trustworthy and verifiable data that increase transparency in identifying pollution sources.

\bibliographystyle{unsrt}
\bibliography{refsa}

\begin{thebibliography}{10}

\bibitem{earthcomp}
Lawrence~Buja Washington, Warren~M. and Anthony Craig.
\newblock The computational future for climate and earth system models: on the
  path to petaflop and beyond.
\newblock {\em Philosophical Transactions of the Royal Society A: Mathematical,
  Physical and Engineering Sciences}, pages 833--846, 2009.

\bibitem{dong2014learning}
Chao Dong, Chen~Change Loy, Kaiming He, and Xiaoou Tang.
\newblock Learning a deep convolutional network for image super-resolution.
\newblock In {\em European conference on computer vision}, pages 184--199.
  Springer, 2014.

\bibitem{Leidig:2017gan}
Christian Ledig, Lucas Theis, Ferenc Huszar, Jose Caballero, Andrew Cunningham,
  Alejandro Acosta, Andrew Aitken, Alykhan Tejani, Johannes Totz, Zehan Wang,
  and Wenzhe Shi.
\newblock {Photo-realistic single image super-resolution using a generative
  adversarial network}.
\newblock {\em Proceedings of the IEEE conference on computer vision and
  pattern recognition}, 1:4681--4690, 2017.

\bibitem{lim2017enhanced}
Bee Lim, Sanghyun Son, Heewon Kim, Seungjun Nah, and Kyoung Mu~Lee.
\newblock Enhanced deep residual networks for single image super-resolution.
\newblock In {\em Proceedings of the IEEE conference on computer vision and
  pattern recognition workshops}, pages 136--144, 2017.

\bibitem{wang2018esrgan}
Xintao Wang, Ke~Yu, Shixiang Wu, Jinjin Gu, Yihao Liu, Chao Dong, Yu~Qiao, and
  Chen Change~Loy.
\newblock Esrgan: Enhanced super-resolution generative adversarial networks.
\newblock In {\em Proceedings of the European Conference on Computer Vision
  (ECCV)}, pages 0--0, 2018.

\bibitem{sentinel5p}
Oliver Schneising, Michael Buchwitz, Maximilian Reuter, Heinrich Bovensmann,
  John Burrows, Tobias Borsdorff, and Nicholas~M. Deutscher.
\newblock A scientific algorithm to simultaneously retrieve carbon monoxide and
  methane from tropomi onboard sentinel-5 precursor.
\newblock {\em Atmospheric Measurement Techniques}, pages 6771--68802, 2019.

\bibitem{kim2016}
Jiwon Kim, Jung~Kwon Lee, and Kyoung~Mu Lee.
\newblock Accurate image super-resolution using very deep convolutional
  networks.
\newblock In {\em Proceedings of the IEEE conference on computer vision and
  pattern recognition}, pages 1646--1654, 2016.

\bibitem{kim2016recursive}
Jiwon Kim, Jung~Kwon Lee, and Kyoung~Mu Lee.
\newblock Deeply-recursive convolutional network for image super-resolution.
\newblock In {\em Proceedings of the IEEE conference on computer vision and
  pattern recognition}, pages 1637--1645, 2016.

\bibitem{dai2019}
Tao Dai, Jianrui Cai, Yongbing Zhang, Shu-Tao Xia, and Lei Zhang.
\newblock Second-order attention network for single image super-resolution.
\newblock In {\em Proceedings of the IEEE conference on computer vision and
  pattern recognition}, pages 11065--11074, 2019.

\bibitem{tai2017}
Ying Tai, Jian Yang, and Xiaoming Liu.
\newblock Image super-resolution via deep recursive residual network.
\newblock In {\em Proceedings of the IEEE conference on computer vision and
  pattern recognition}, pages 3147--3155, 2017.

\bibitem{tai2017memnet}
Ying Tai, Jian Yang, Xiaoming Liu, and Chunyan Xu.
\newblock Memnet: A persistent memory network for image restoration.
\newblock In {\em Proceedings of the International Conference on Computer
  Vision}, pages 4539--4547, 2017.

\bibitem{gan}
Ian Goodfellow, Jean Pouget-Abadie, Mehdi Mirza, Bing Xu, David Warde-Farley,
  Sherjil Ozair, Aaron Courville, and Yoshua Bengio.
\newblock Generative adversarial nets.
\newblock In {\em Advances in neural information processing systems}, pages
  2672--2680, 2014.

\bibitem{fukami2018super}
Kai Fukami, Koji Fukagata, and Kunihiko Taira.
\newblock Super-resolution reconstruction of turbulent flows with machine
  learning.
\newblock {\em arXiv preprint arXiv:1811.11328}, 2018.

\bibitem{jiang2020meshfreeflownet}
Chiyu~Max Jiang, Soheil Esmaeilzadeh, Kamyar Azizzadenesheli, Karthik
  Kashinath, Mustafa Mustafa, Hamdi~A Tchelepi, Philip Marcus, Anima
  Anandkumar, et~al.
\newblock Meshfreeflownet: A physics-constrained deep continuous space-time
  super-resolution framework.
\newblock {\em arXiv preprint arXiv:2005.01463}, 2020.

\bibitem{bode2019using}
Mathis Bode, Michael Gauding, Zeyu Lian, Dominik Denker, Marco Davidovic,
  Konstantin Kleinheinz, Jenia Jitsev, and Heinz Pitsch.
\newblock Using physics-informed super-resolution generative adversarial
  networks for subgrid modeling in turbulent reactive flows.
\newblock {\em arXiv preprint arXiv:1911.11380}, 2019.

\bibitem{div2k}
Eirikur Agustsson and Radu Timofte.
\newblock Ntire 2017 challenge on single image super-resolution: Dataset and
  study.
\newblock In {\em Proceedings of the IEEE Conference on Computer Vision and
  Pattern Recognition Workshops}, pages 126--135, 2017.

\bibitem{flickr2k}
Radu Timofte, Eirikur Agustsson, Luc Van~Gool, Ming-Hsuan Yang, and Lei Zhang.
\newblock Ntire 2017 challenge on single image super-resolution: Methods and
  results.
\newblock In {\em Proceedings of the IEEE conference on computer vision and
  pattern recognition workshops}, pages 114--125, 2017.

\bibitem{urban100}
Jia-Bin Huang, Abhishek Singh, and Narendra Ahuja.
\newblock Single image super-resolution from transformed self-exemplars.
\newblock In {\em Proceedings of the IEEE conference on computer vision and
  pattern recognition}, pages 5197--5206, 2015.

\bibitem{Loffler:2011ay}
Frank Löffler, Joshua Faber, Eloisa Bentivegna, Tanja Bode, Peter Diener,
  Roland Haas, Ian Hinder, Bruno~C. Mundim, Christian~D. Ott, Erik Schnetter,
  Gabrielle Allen, Manuela Campanelli, and Pablo Laguna.
\newblock {The Einstein Toolkit: A Community Computational Infrastructure for
  Relativistic Astrophysics}.
\newblock {\em Class. Quant. Grav.}, 29:115001, 2012.

\bibitem{kranc}
Sascha Husa, Ian Hinder, and Christiane Lechner.
\newblock {Kranc: a Mathematica package to generate numerical codes for
  tensorial evolution equations}.
\newblock {\em Computer Physics Communications}, 174:983--1004, 2006.

\bibitem{2004CQGra..21.1465S}
Erik Schnetter, Scott~H. Hawley, and Ian Hawke.
\newblock {Evolutions in 3D numerical relativity using fixed mesh refinement}.
\newblock {\em Classical and Quantum Gravity}, 21(6):1465--1488, March 2004.

\end{thebibliography}

\clearpage
\appendix
\begin{appendices}

\section{Simulation}
\label{sec:app}

To integrate the advection-diffusion equation  (\ref{eq:ade}), a numerical code is leveraged generated with the Cactus 
framework~\cite{Loffler:2011ay}, using the automated code-generation package 
Kranc~\cite{kranc} and the Carpet mesh-refinement module~\cite{2004CQGra..21.1465S}.
A fourth-order finite differencing for spatial derivatives, and the Method of Lines 
technique (combined with fourth-order Runge-Kutta integration) is used to advance the solution in
time. Periodic boundary conditions are applied to the outer boundaries.

In order to solve Equation~(\ref{eq:ade}) for gas concentration $C$, we first need to model the velocity $\bf{u}$ and source $S$ fields.
$S$ is modeled as a piece-wise constant function of compact spatial support (typically, as a function which is zero everywhere, except for a few scattered circles where $S$ intermittently
switches on and off).





In the advection-diffusion equation, the velocity field {\bf u} is assumed to be spatially constant with fluctuations in time around a given direction (i.e., the $x$ direction):  
    \begin{equation}
        {\bf u}(t) = \{u_0 + \sum_{i=1}^T{u^x_i \cos(k^x_i t + \phi^x_i)}, \sum_{i=1}^T{u^y_i \cos(k^y_i t + \phi^y_i)}\} \qquad u^x_i,u^y_i \ll u_0
    \end{equation}
  
\begin{figure}[htbp]
    \centering
    \includegraphics[width=\linewidth]{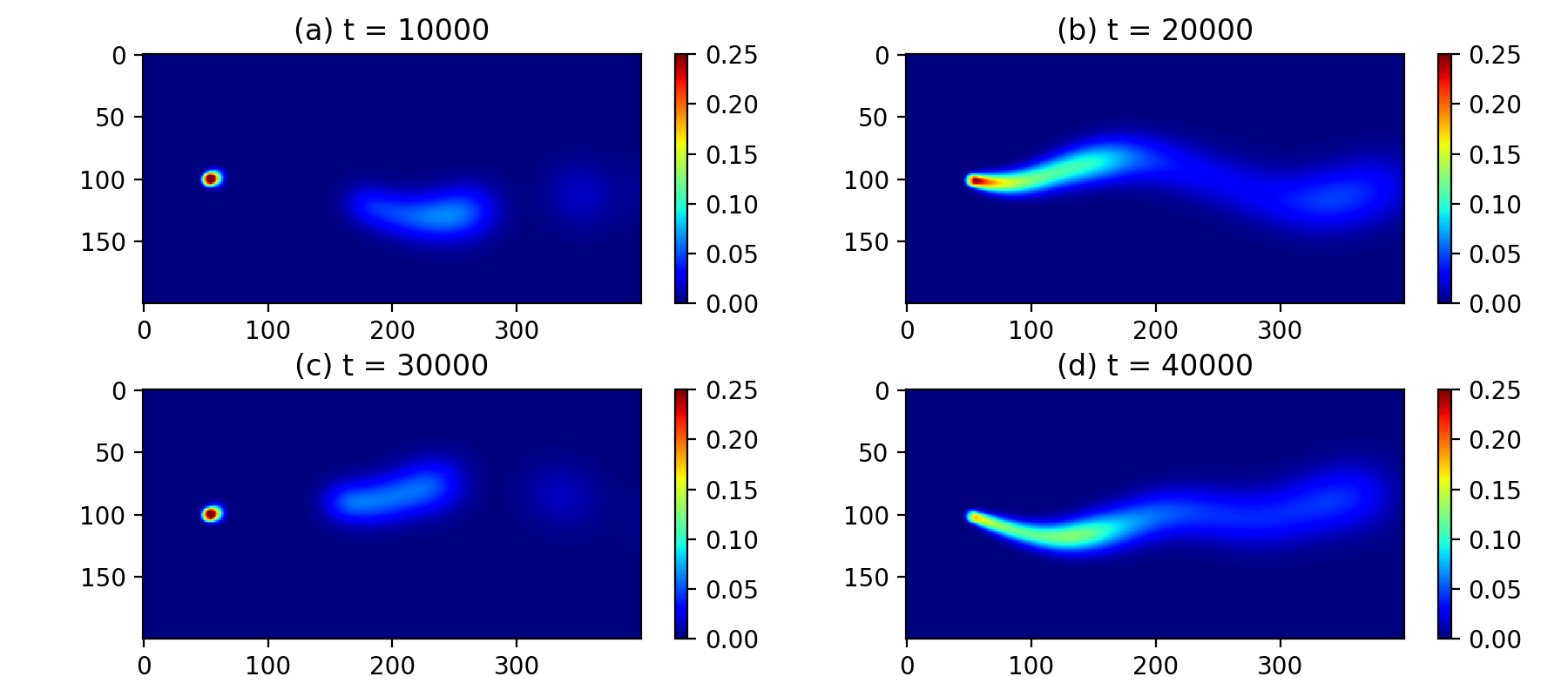}
    \caption{
        Snapshots from the evolution of a single plume model at different iteration steps.
    }
    \label{fig:sims}
\end{figure}

Figure~\ref{fig:sims} shows snapshots from a single-source plume model. Multiple-source profiles are obtained by superimposing several single-source solutions, centered at different points in the simulation domain. Since Equation~(\ref{eq:ade}) is linear in $C$ and the wind velocity is spatially constant, the resulting superposition satisfies Equation~(\ref{eq:ade}) if so do the single-source components. 


\section{Training details}
\label{sec:train}

The LR simulations are of size $100 \times 50$ 
and HR simulations are of size $400 \times 200.$ 
Since the simulation model is expressed by a linear differential equation, an arbitrary superposition of the concentration map also satisfies the physics model. Moreover, the periodicity in wind velocity makes it possible to align concentration at different iteration steps as long as they differ by an integer number of periods. In particular, each training image has 20 plume sources randomly placed within the frame. The source flux is randomly sampled from a uniform distribution between 0 and maximum flux. Datasets are stored as 2D images with floating-point format ranging between 0 and 1. In total there are 2000 images with different random source placements, intensities, and start times.

At each training step, $16 \times 16$ LR patches are used as input and the corresponding HR patch size is $64 \times 64$. For all the experiments we empirically set the weight $\eta = 100.0$ to balance the ratio between the physics consistency loss and the pixel loss. A Cosine Annealing learning rate scheduler with restarts is used, adjusting the learning rate to decay from $2\times 10^{-4}$ to $1\times 10^{-7}$ within $2.5\times 10^5$ iterations, and the whole process repeats 4 times. Adam optimizer with $\beta_1 = 0.9$ and $\beta_2 = 0.999$ is used for optimization. The model is implemented in PyTorch. 

\section{Additional results}
\label{sec:add_result}

\begin{figure}[htbp]
    \centering
    \includegraphics[width=\linewidth]{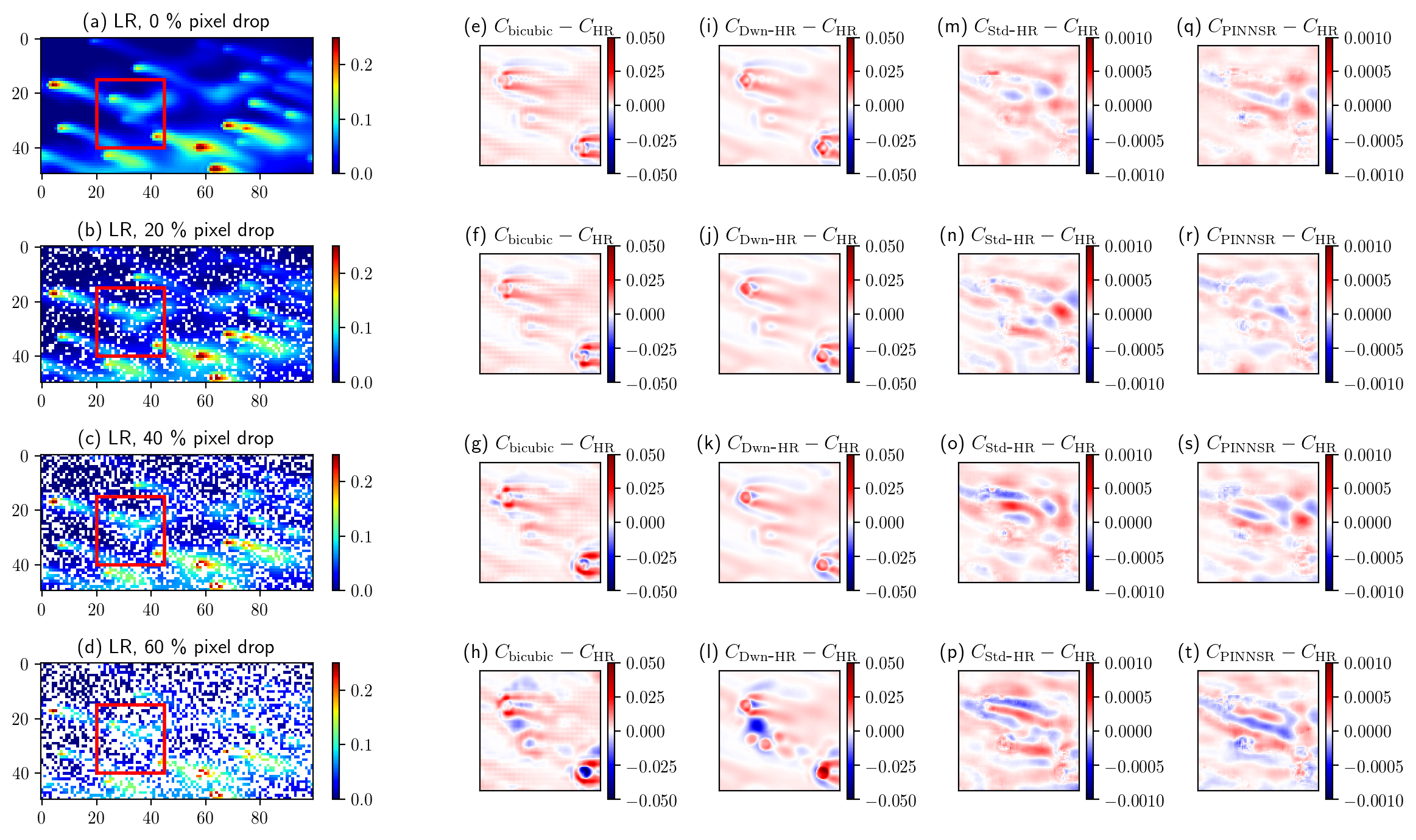}
    \caption{
        Qualitative comparison between different SR models for all pixel dropping rates. (a) - (d) LR input for 0\% to 60\% pixel drop rate; (e) - (h) SR generated by bicubic; (i) - (l) SR generated by Dwn-HR; (m) - (p) SR generated by Std-SR; (q) - (t) SR generated by PINNSR. Each row has the same pixel drop percentage. 
    }
    \label{fig:vis_all}
\end{figure}

For four scenarios where 0\%, 20\%, 40\% and 60\% of the pixels are dropped, 
the visual results are shown in Figure~\ref{fig:vis_all}. Firstly, a higher missing pixel percentage corresponds to a larger pixel loss, and thus lowers PSNR. Secondly, within the same pixel drop percentage, PINNSR $<$ Std-SR $\ll$ bicubic $\approx$ Dwn-HR in terms of pixel error. This is also consistent with the results reported in Table~\ref{table:compare}.

\section{Down-sampled HR vs. native LR input}
\label{sec:down-sample}

Traditionally, the input of the SR NN is generated by down-sampling the HR. However, in this work, we emphasize using native LR simulation as input. Here we demonstrate that down-sampled HR as input is an oversimplification of the problem.

Following the traditional process, let's consider the NN trained with down-sampled HR as input. The results are shown in Table~\ref{table:compare_native}. When the NN is also tested with down-sampled HR as input, it would result in PSNR as high as $\sim 100$. However, this does not reflect the actual performance. At test time, the HR ground truth is not available, only native LR is available as input. As shown in column 2 of Table~\ref{table:compare_native}, the PSNR is $\sim 45$, which is only on par with the performance of the bicubic baseline.

\begin{table}[htbp]
  \caption{Evaluation for models trained with down-sampled HR as LR input. All models have poor performance on reconstructing native LR.}
  \label{table:compare_native}
  \centering
  \begin{tabular}{ccc}
    \toprule
            & \multicolumn{2}{c}{$\rm PSNR$ / $1 - \rm SSIM$ / 
            $\mathcal{L}_{\rm phys}$}      \\
    \cmidrule(r){2-3}
    Pixel drop   & Test with down-sampled HR as input & Test with native LR as input  \\
    \midrule
    0\%     & 100.49 / 0.0 / 6.4E-8 & 
              45.85 / 0.0058 / 1.1E-6  \\
              
    20\%    & 98.03 / 6E-8 / 6.6E-8 & 
              45.17 / 0.0064 / 1.5E-6 \\
              
    40\%    & 98.69 / 6E-8 / 7.0E-8 & 
              44.83 / 0.0065 / 1.1E-6 \\
              
    60\%    & 98.49 / 0.0 / 13.1E-8 & 
              44.66 / 0.0061 / 0.8E-6\\
    \bottomrule
  \end{tabular}
\end{table}

\section{Bicubic interpolation for missing pixel reconstruction}
\label{sec:bicubic_interp}

It is also possible to leverage bicubic interpolation to reconstruct missing pixels, and subsequently use the 0\% pixel drop PINNSR model for upsampling. The results are summarized in Table~\ref{table:bic_interpolation}. Recovering missing pixels using bicubic interpolation is slightly better than the bicubic baseline, but it does not have comparable performance to the models directly trained with missing pixel LR (Table~\ref{table:compare}).

\begin{table}[htbp]
  \caption{Evaluation for bicubic interpolation as a preprocessing step with 0\% missing pixel NN model. The performance is not comparable to the models trained with missing pixels.}
  \label{table:bic_interpolation}
  \centering
  \begin{tabular}{ccc}
    \toprule
            & \multicolumn{2}{c}{$\rm PSNR$ / $1 - \rm SSIM$ / 
            $\mathcal{L}_{\rm phys}$}      \\
    \cmidrule(r){2-3}
    Pixel drop   & Std-SR (0 \%) & PINNSR (0 \%) \\
    \midrule
    0\%     & 82.29 / 2.1E-6 / 2.8E-7 & 
              82.83 / 2.0E-6 / 0.9E-7 \\
              
    20\%    & 56.68 / 6.8E-4 / 3.6E-6 & 
              57.16 / 6.6E-4 / 2.9E-8 \\
              
    40\%    & 51.67 / 0.0025 / 4.4E-6 & 
              51.84 / 0.0024 / 3.8E-6 \\
              
    60\%    & 46.90 / 0.0076 / 2.7E-6 & 
              46.92 / 0.0080 / 2.7E-6 \\
    \bottomrule
  \end{tabular}
\end{table}

\end{appendices}

\end{document}